# Integrating Knowledge Bases and Statistics in MT

Kevin Knight, Ishwar Chander, Matthew Haines,
Vasileios Hatzivassiloglou, Eduard Hovy, Masayo Iida, Steve K. Luk,
Akitoshi Okumura, Richard Whitney, Kenji Yamada

USC/Information Sciences Institute
4676 Admiralty Way
Marina del Rey, CA 90292
{knight,chander,haines,vh,hovy,iida,luk,okumura,whitney,kyamada}@isi.edu

## Abstract

We summarize recent machine translation (MT) research at the Information Sciences Institute of USC, and we describe its application to the development of a Japanese-English newspaper MT system. Our work aims at scaling up grammar-based, knowledge-based MT techniques. This scale-up involves the use of statistical methods, both in acquiring effective knowledge resources and in making reasonable linguistic choices in the face of knowledge gaps.

## 1 Goals

Knowledge-based machine translation (KBMT) techniques have yielded high quality MT systems in narrow problem domains (Nirenburg *et al.* 1992; Nyberg & Mitamura 1992). This high quality is delivered by algorithms and resources that permit some access to the meaning of texts. But can KBMT be scaled up to unrestricted newspaper articles? We believe it can, provided we address two additional questions:

1. In constructing a KBMT system, how can we acquire knowledge resources (lexical, grammatical, conceptual) on a large scale?

2. In applying a KBMT system, what do we do when definitive knowledge is missing?

There are many approaches to these questions. Our working hypotheses are (1) a great deal of useful knowledge can be extracted from online dictionaries and text; and (2) statistical methods, properly integrated, can effectively fill knowledge gaps until better knowledge bases or linguistic theories arrive. This paper describes completed and ongoing research on these hypotheses. This research is tightly coupled with our development effort on a large-scale Japanese-English MT system, as part of the ARPA-sponsored PANGLOSS project (NMSU/CRL, USC/ISI, & CMU/CMT 1994; Nirenburg & Frederking 1994; Knight & Luk 1994). Sample translations from our system appear in Section 4.

## 2 System Design: Philosophy

At an abstract level, all MT systems are composed of two types of components: transformers (T) and ranker/pruners (R). A T-component takes some input structure and outputs zero or more new structures. The inputs and outputs may be of the same type (e.g., string → string) or different types (e.g., string → parse tree). An R-component takes a set of structures of the same type, assigns each structure a score, then prunes away some number of low-scoring structures.

For example, the statistical MT system Candide (Brown *et al.* 1993) operates in a T-R-R fashion. A French string is transformed into many thousands of French-English string pairs. These translations are ranked and pruned first by a translation model, then by an English-only language model. Both rankers are "soft", never assigning a zero score. The final pruning leaves only a single translation.

The LINGSTAT system (Yamron *et al.* 1994), works with a T-R-T-R design. The first T-component transforms a Japanese sentence into many possible parse trees, as stipulated by a context-free grammar. A statistical ranker chooses one "best" tree and prunes the rest. A second T-component turns that tree into a set of possible English translations, using a bilingual dictionary and word ordering rules. These translations are ranked by a language model similar to the one employed by Candide.

KBMT systems typically use a T-T-R-T sequence. First, the source text is turned into a set of syntactic analyses. These analyses are typically not ranked until a second transformation turns them into a set of candidate semantic representations. A "hard" ranker then prunes away any structure containing a semantic constraint violation (according to a database of such constraints). With luck, exactly one candidate survives. A final T-component turns it into a target language string. In practice, such generators produce one output per input, so no further ranking is necessary.

Our KBMT design is a variant of this one. Unfortunately, in translating newspaper text, we cannot rely on a small vocabulary, a tightly controlled input grammar, or a perfect domain model for semantic ranking. We must scale up along several dimensions at once, the result being that knowledge gaps are inevitable. To fill in these gaps, we have made several design changes. One change is that our semantic ranker is "soft". It assigns a non-zero score to every semantic candidate, based on the number and seriousness of the constraint violations it contains. We also employ statistical methods that enable us to handle ungrammatical input, name translation, incomplete interlingual analyses, etc. At every stage in the translation path, we have had to ask ourselves: What is the most appropriate technique? What are the tradeoffs among accuracy, robustness, efficiency, and human effort? Our solutions are described in following sections, on a module-by-module basis.

Our only real philosophical departure from standard KBMT practice is that we remain flexible about which components should handle which linguistic phenomena. For example, if our semantic representations do not contain enough stylistic information to allow an intelligent choice between two expressions, we may simply let a target-language statistical model choose. In the extreme, we can even skip semantics altogether (see Section 3.10), forcing problems like word sense disambiguation to be resolved downstream. But our longer-term goal is to provide timely knowledge to each component.

## 3 System Design: Modules and Knowledge Resources

Our system building efforts began in January 1994, preceded by several months of collecting dictionaries and corpora and setting up a Japanese computational environment. We set our first goal to be high-quality, human-assisted translation of a single text; we then added components and preference knowledge to remove the human. We then set about improving coverage and robustness.

Figure 1 illustrates the current Japanese-English MT system, including all processing modules and knowledge resources. The rest of this section covers these systems in detail.

### 3.1 Segmenter/Tagger (JUMAN)

Japanese writing uses no inter-word spacing, so our first step in translation is to decide on word breaks. We use the JUMAN preprocessor (from Kyoto University) for this task. JUMAN has a large dictionary (on the order of 100,000 words) as well as combination rules that perform part-of-speech tagging.

### 3.2 Chunker

We originally intended to send JUMAN output directly to the parser, but we found that a few intermediate steps could substantially reduce both JUMAN error and syntactic ambiguity. Our current chunker performs four tasks: (1) dictionary-based word chunking (compounds found in a CD-ROM dictionary are resegmented), (2) recognition of personal, corporate and place names, (3) number recognition and resegmenting, and (4) phrase chunking. This last type of chunking inserts new strings into the input sentence, such as "BEGIN-NP" and "END-NP". The syntactic grammar is written to take advantage of these barriers, prohibiting the combination of partial constituents across phrase boundaries. VP and S chunking allow us to process the very long sentences characteristic of newspaper text, and to reduce the number of syntactic analyses.

All chunking is driven by patterns that operate like finite automata. Sample rules are:

```
(TOPIC-HA      (N+ HA DATE COMMA ^))
(TOPIC-HA      (N+ HA DATE KARA COMMA ^))
(VP-BEGIN   == (is TOPIC-LEAD))
(VP-BEGIN   == (is Q-VP-BEGIN))
(VP-END     == (is PRD))
(VP            (VP-BEGIN < ANY1+ > VP-END) :left <<VP :right VP>>)
```

### 3.3 Parser (HAX)

We made an early decision to do all syntactic and semantic processing bottom-up, to take advantage of partial analyses. We designed and built HAX, a bottom-up chart parser. HAX parses according to a context-free grammar augmented with unification feature equations, in the style of (Shieber 1986). HAX grammars may contain complex disjunctions of equations, exclusive-or disjunctions, feature existence checking, and simple negation. Our current grammar of Japanese contains 71 unary rules, 32 binary rules, and 15 n-ary rules. A typical rule looks like:

```
((NP -> S NP)
   ((X1 syn infl) = (*OR* kihon ta-form rentai))
   ((X0 syn) = (X2 syn))
   ((X0 syn comp) = plus)
   ((X0 syn s-mod) = (X1 syn)))
```

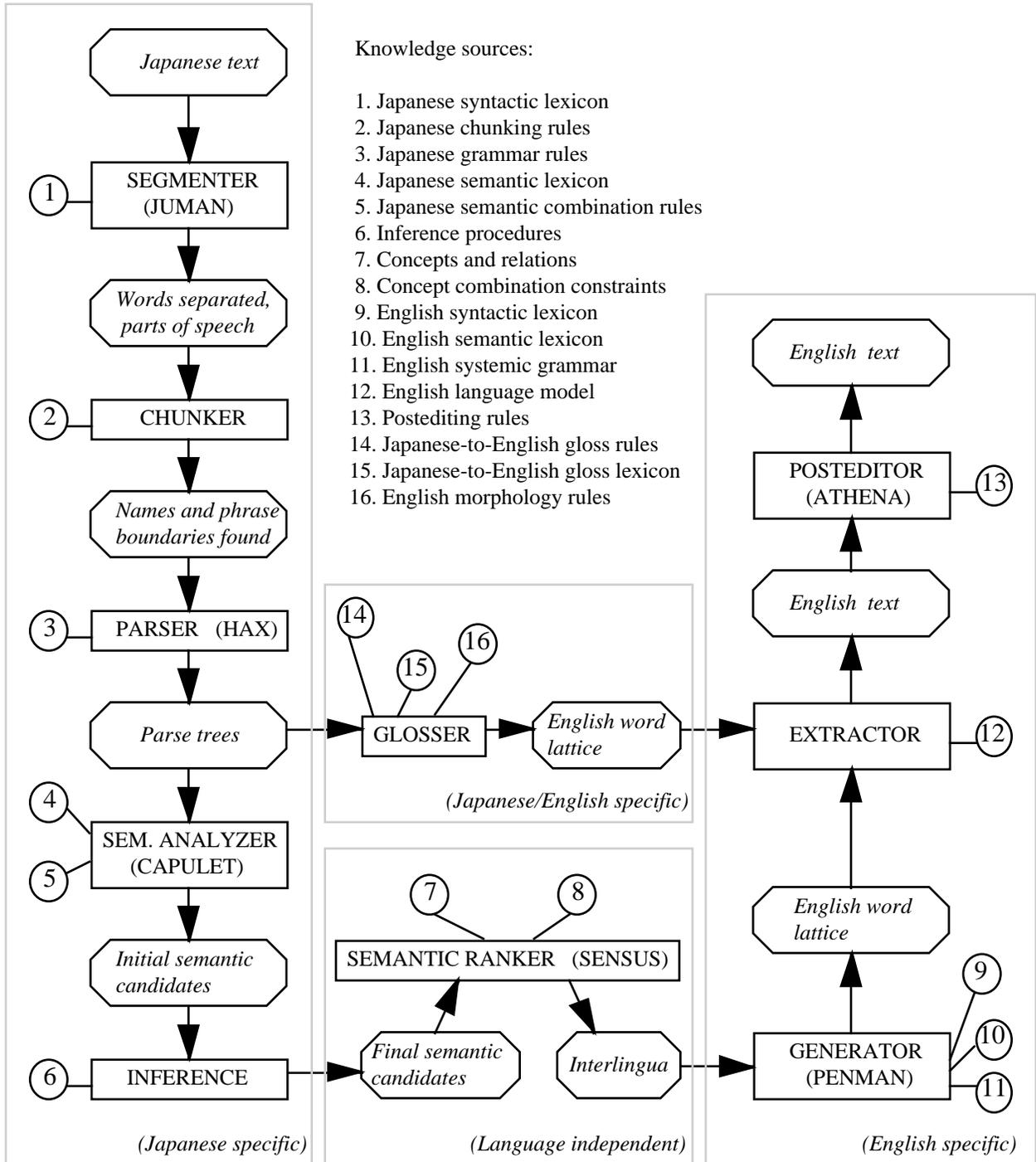

Figure 1: Design for the USC/ISI Japanese-English MT system.

This rule builds a noun phrase (X0) from a relative clause (X1) and another noun phrase (X2). It also tests an inflectional feature of the clause, inherits syntactic features from the child noun phrase to the parent, and adds new features.

HAX outputs a parse forest, represented as a list of constituents with features that were assigned to them during parsing. A full parse may or may not be found. We strive for full parses whenever possible, because we cannot rely on deep semantics to patch everything up. Since one unanticipated word or punctuation mark may splinter the input into four or five pieces, we are working on techniques for word-skipping, inspired in part by (Lavie 1994).

> *Ongoing research project.* We are investigating methods to repair trouble spots in grammatical analysis. In particular, we are looking at statistical differences between fully-parsed and not-fully-parsed sentences. These differences include relative distributions of part-of-speech bigrams. Our goal is to provide automatic feedback for grammar development and to identify word skipping or tag repair possibilities for exploitation at runtime.

### 3.4 Semantic Analyzer (CAPULET)

Like our parser HAX, the semantic analyzer CAPULET operates bottom-up. It assigns zero or more semantic interpretations to each syntactic constituent in the HAX parse forest. At the bottom, word meanings are retrieved from a semantic lexicon. Moving up the parse trees, meanings of constituents are computed compositionally from the meanings of their children, according to a database of semantic combination rules. Semantic rules have the same format as syntactic rules; they are keyed to one another via their context-free parts, in the style of (Dowty, Wall, & Peters 1981) and (Moore 1989). Semantic rules look like:

```
((NP -> S NP)
   ((X2 syn form) = (*NOT* rentaidome))
   ((X0 sem instance) = rc-modified-object)
   ((X0 sem head) = (X2 sem))
   ((X0 sem rel-mod) = (X1 sem))
   (*OR* (((X1 map subject-role) =c X2))
         (((X1 map object-role)  =c X2))
         (((X1 map object2-role) =c X2))))
```

The current semantic rule base contains at least one rule for every syntactic rule in the grammar.

The output of semantic analysis is represented in a number of ways. Inside the analyzer, semantic information takes the form of directed acyclic graphs. For semantic ranking, we view the graphs as lists of assertions, as in (Hobbs 1985). We also use the SPL format of the PENMAN generation system (Penman 1989). Here is a sample output, representing the sentence "the new company plans to launch in February":

```
(|h-709| / |have as a goal|
   :SENSER (|c-710| / |company/business|
              :Q-MOD (|n-711| / |new~virgin|))
   :PHENOMENON (|f-712| / |found, launch|
                  :TEMPORAL-LOCATING (|c-713| / |calendar month| :MONTH-INDEX 2)
                  :AGENT |c-710|)
   :THEME |c-710|)
```

The acquisition of a large Japanese semantic lexicon is a difficult task. We are tackling this problem with a combination of automatic and manual techniques. (Okumura & Hovy 1994) and (Knight & Luk 1994) describe algorithms for using bilingual dictionaries to propose links between non-English words and concepts in a knowledge base. We have also built an interface called the Acquisitor, which allows a person to identify word-concept links by conceptually tagging words in context. This provides a distribution of senses for each word and potentially offers data for topic-based disambiguation algorithms. This acquisition work requires substantial resources; at present we are working with five part-time knowledge enterers ("acquisitors").

Another issue is handling unknown words, especially technical terms and proper names. This is a good-news/bad-news situation for Japanese-English translation. The bad news is that many names and terms are written in non-Roman Japanese orthography. We cannot simply translate X as X, as we might in Spanish-English translation. The good news is that large numbers of terms and names are loan words from English, and are written in the phonetic katakana script. The problem is to guess which English (or English-looking) word was the most probable source for a given katakana transliteration.

> *Ongoing research project.* We are investigating the use of speech recognition techniques in translating katakana words into English. These techniques include acoustic matching for known English words and statistical noisy channel models for unknown words. Our goal is to produce an accurate Katakana-English transliterator.

### 3.5 Inference

Semantic analysis captures information explicit in the source text and does not, for the most part, include implicit information necessary for target language generation. The job of the inference module is to make that implicit material explicit. Currently, inference is the least developed part of the working system; its only jobs are to reorganize relative clauses and to insert topical items into semantic roles.

### 3.6 Semantic Ranking (SENSUS)

Semantic analysis has now given us a set of candidate meanings, in terms of concepts and relations from our knowledge base. This inventory of concepts is a synthesis of resources like the PENMAN Upper Model (Bateman 1990), ONTOS (Carlson & Nirenburg 1990), the WordNet thesaurus (Miller 1990), and the Longman Dictionary of Contemporary English (Longman-Group 1978). It contains roughly 70,000 terms organized into inheritance networks; (Knight & Luk 1994) describe its construction.

The concept inventory and its functional interface are part of a system called SENSUS. The other part is a set of constraints on which concepts are naturally related to which others in our world. Thus, SENSUS is a world model, (mostly) independent of particular natural languages. It can be used to rank candidate meanings produced by analysis of English, Japanese, Spanish, etc. The model scores meanings similar to the way a trigram word model scores sentences. We consider all conceptual "trigrams" in the semantic representation (e.g., `<EAT-2, PATIENT, WORM-1>`, `<EAT-2, AGENT, PERSON-1>`, ...) and assign to each a probability, based on considerations like domain/range constraints on relations and mutual disjointness of classes. The trigram scores are multiplied together to yield an overall measure of semantic coherence. As with word trigrams, we smooth so that no interpretation is assigned a zero score.

The main issue regarding semantic constraints is one of large-scale acquisition. Again, we are pursuing both manual and automatic approaches. We have developed a constraint language for manually entered world knowledge. This language takes advantage of the SENSUS hierarchy, so that a single rule can cover a very large number of cases. The language allows for relaxed constraints (Carlson & Nirenburg 1990), giving us a softer measure of violation.

> *Ongoing research project.* We are looking at ways to automatically identify semantic constraints using machine learning techniques. We are using untagged and concept-tagged corpora (single and dual-language) and the prior knowledge of SENSUS to learn which concepts co-occur in which configurations. Our goal is a system that can propose semantic constraints that improve the performance of translation.

The output of the semantic ranker is the same as the input, except that each candidate meaning is assigned a score (as described earlier). The highest scoring meaning is called the interlingual representation.

## 3.7 Generator (PENMAN)

PENMAN is a natural language sentence generation program in development at USC/ISI since 1982. It provides computational technology for generating English sentences, starting with input specifications of a non-linguistic kind. The culmination of a continuous research effort since 1978, Penman embodies one of the most comprehensive computational generators of English sentences in the world. A detailed overview of language generation in Penman can be found in (Penman 1989); for a description of PENMAN's use in the PANGLOSS MT system, see (NMSU/CRL, USC/ISI, & CMU/CMT 1994).

PENMAN converts an interlingua expression into English. For example, it turns the analysis two pages ago into:

```
The company plans the launching in February.
```

PENMAN is powered by a large grammar of English, and by a large morpho-syntactic lexicon extracted from online resources and linked to concepts in SENSUS. This lexicon contains about 91,000 root forms, including many phrasal verbs.

When PENMAN does not have enough knowledge to generate a sentence, or its input is incomplete, it falls back on heuristics and defaults. For example, if the interlingua does not specify definiteness and time, PENMAN will generate the article *the* and use present tense. PENMAN also uses the SENSUS hierarchy to make judgments about which syntactic structures to generate, e.g., leveraging off the fact that semantically similar verbs often have similar syntactic behaviors.

Our research on generation is aimed at improving aspects of PENMAN, e.g., by repairing cases where the defaults and heuristics produce an incorrect result. Our approach is to generate many possible sentences packed into a space-efficient word lattice, and have a ranker/pruner choose the best one. For example, we might not be sure if "in Japan" or "on Japan" or "at Japan" is a proper way to denote location in English. But a statistical n-gram language model has a clear preference. The same model can prefer "the moon" to "a moon". We can therefore address, if not solve, a variety of problems with a single method. Of course, definitive knowledge always serves to narrow the range of choices presented to the ranker. If we ever achieved perfect knowledge of English generation, the word lattice would degenerate to a definitive single path, and no ranking or statistics would be needed.

> *Ongoing research project.* We are designing a many-paths generator to address problems with lexical choice, non-semantic collocation generation, incomplete lexical specification, and incomplete interlingua input.

## 3.8 Extractor

We have built several programs for statistical language modeling. The main runtime program accepts a word lattice from the generator (or glosser—see Section 3.10) and extracts the most likely path through that lattice. Likelihood is based on a word trigram model we built from online Wall Street Journal text. We use a version of enhanced Good-Turing smoothing (Church & Gale 1991) to assign non-zero probabilities to previously unseen trigrams. This form of smoothing is detailed enough to distinguish "good" unseen trigrams (e.g., *joyful lovely days*) from "bad" ones, even those composed of popular unigrams (e.g., *the an the*). Our current experiments involve re-ordering the top N paths found by the extractor, using information about longer-distance collocations.

## 3.9 Automated Posteditor (ATHENA)

The final stage of translation serves to fix some of the problems unaddressed or poorly addressed in previous modules. The posteditor is capable of basic string matching and transformation for local repairs, but it also has a special ability to insert English articles (*a*, *an*, *the*) into article-free text. This is important for us, because Japanese noun phrases are rarely marked for definiteness. The posteditor's knowledge is statistical, but its model is deeper than that of the trigram model discussed above. It organizes its knowledge into automatically generated decision trees that insert articles based on features of the surrounding context; (Knight & Chander 1994) has details.

> *Ongoing research project.* We are expanding our statistical posteditor to predict/insert new articles (*some*, *any*, and the *null* article), plural morphology, and prepositions.

### 3.10 Glosser

At the top of this paper, we described our strategy for filling knowledge gaps with statistical methods. In the extreme case, we suggested the option of bypassing semantic processing altogether. The glosser is a program that does just this, by transforming a Japanese parse forest directly into an English word lattice. (The glosser is shown in the center of Figure 1.)

The glosser itself is nearly identical to the semantic analyzer, CAPULET. It traverses the parse trees bottom-up, but instead of assigning and composing meanings at each constituent, it assigns and concatenates target language strings. Japanese lexical items are assigned English strings distilled from several large Japanese-English dictionaries (160,000 entries, covering about 95% of the words in our texts). These strings are concatenated, and new words are inserted, as the glosser moves up the tree.

Strings themselves are represented as feature structures, e.g.:

```
((gloss ((op1 "John")
         (op2 ((op1 (*or* "wants" "want"))
               (op2 ((op1 "to")
                     (op2 "eat")))))
         (op3 "now"))))
```

A unique feature of the glosser is its exploitation of feature unification. Features can represent not only strings, but also more abstract characteristics of the input. For example, we can set `tmp` features like `((passive +))` and `((past +))` as we accumulate Japanese verb suffixes, delaying the production of an English word lattice until we have a complete description of the verb complex. This use of features allows us to escape strict compositional concatenation of strings. Here is a sample glosser rule:

```
((NP -> S NP)
  ((X0 gloss op1) = (X2 gloss))
  ((X0 gloss op2) = (*or* "which" "that"))
  ((X0 gloss op3) = (X1 gloss))
  ((X0 tmp) = (X2 tmp)))
```

While CAPULET's output is ranked and pruned by SENSUS, the glosser's output goes directly to the trigram-based extractor described above.

The glosser-based MT system can translate any Japanese text. It does not achieve high-quality, but getting the glosser running was a good rehearsal for our upcoming integration of a large semantic lexicon, preference rules, etc. It also forced us to address problems with unknown words, missing grammar rules, overchunking, and wrong segmentation.

## 4 Conclusion and Sample Translations

This paper has reported progress on a knowledge-based machine translation system for Japanese-English being developed at USC/ISI. With small, hand-crafted knowledge bases, the system translates a few texts at high quality. With no semantic knowledge, the system translates any text at lower quality. Our research aims to close the gap, increasing the quality of newspaper text translation by providing more knowledge as well as robustness in the face of missing knowledge.

### 4.1 Interlingua-Based Output Sample

```
Citizen Watch announced on the eighteenth to establish a joint venture with Perkin Elmer
Co.  a major microcircuit manufacturing device manufacturer and to start the production
of the microcircuit manufacturing device.
  The new company plans a launching in February.
  The subsidiary of Perkin Elmer Co.  in Japan bears a majority of the stock, and the
production of the dry etching device that is used for the manufacturing of the
microcircuit chip and the stepper is planned.
```

### 4.2 Glosser-Based Output Sample

```
The "gun possession" penalties important--national police agency plans of gun sword legal
reform.
```

```
  The national police agency defend policies that change some of gun sword law that put
the brakes on the proliferation of the 31st gun.  Change plans control recovery plan of
the wrong owning step currency--make the decision by the cabinet meeting of the middle of
this month in three pieces support estimate of the filing in this parliament.
```

## 5 Acknowledgments

Thanks to Licheng Zeng for help in knowledge base construction, and to Oleg Kapanadze and Nadia Mesli for discussions and improvements to our grammar formalism. This work is being carried out under ARPA Order No. 8073, contract MDA904-91-C-5224, and with support from the Department of Defense.